\documentclass{article}[11pt]

\usepackage{microtype}
\usepackage{graphicx}
\usepackage{subfigure}
\usepackage{booktabs} 
\usepackage{verbatim}
\usepackage{enumitem}
\setlist[1]{itemsep=-5pt}
\setlist[itemize]{noitemsep, topsep=0pt}
\usepackage{amsfonts,dsfont}
\usepackage{multirow}
\usepackage{caption} 
\usepackage{varwidth}
\usepackage{xcolor}
\usepackage{amsmath}

\usepackage{hyperref}

\newcommand{\Popu}{X} 
\newcommand{\Indv}{x}

\newcommand{\SetOutcomes}{Y} 
\newcommand{\Outcome}{y}
\newcommand{\predictor}{\mathcal{H}}
\newcommand{\protAttr}{A}
\newcommand{\unprotAttr}{Z}
\newcommand{\bP}{\mathbb{P}}

\newcommand\given[1][]{\:#1\vert\:}

\newcommand{\IndDistanceMetric}{d}
\newcommand{\DistDistanceMetric}{D}
\newcommand{\DistOutcomes}{\mathcal{H}}

\newcommand{\Benefit}{\mathbb{B}}
\newcommand{\all}{\textnormal{ for all }}
\makeatletter
\newcommand*{\balancecolsandclearpage}{%
  \close@column@grid
  \clearpage
  \twocolumngrid
}

\makeatletter
\newcommand*{\rom}[1]{\expandafter\@slowromancap\romannumeral #1@}
\makeatother

\newtheorem{myDefinition}{Definition}

\usepackage[accepted]{icml2018}
\setcitestyle{numbers}

\icmltitlerunning{On Formalizing Fairness in Prediction with Machine Learning}

\begin{document}

\twocolumn[
\icmltitle{On Formalizing Fairness in Prediction with Machine Learning}



\icmlsetsymbol{equal}{*}

\begin{icmlauthorlist}
\icmlauthor{Pratik Gajane}{leo}
\icmlauthor{Mykola Pechenizkiy}{tue}
\end{icmlauthorlist}

\icmlaffiliation{leo}{Department of Computer Science, Montanuniversitat Leoben, Austria, pratik.gajane@unileoben.ac.at}
\icmlaffiliation{tue}{Department of Computer Science, TU Eindhoven, the Netherlands, m.pechenizkiy@tue.nl}

\icmlcorrespondingauthor{Pratik Gajane}{pratik.gajane@unileoben.ac.at }

\icmlkeywords{fairness, fairness-measures, discrimination, machine learning, survey}

\vskip 0.3in
]



\printAffiliationsAndNotice{}  

\begin{abstract}
Machine learning algorithms for prediction are increasingly being used in critical decisions affecting human lives. Various fairness formalizations, with no firm consensus yet, are employed to prevent such algorithms from systematically discriminating against people based on certain attributes protected by law. The aim of this article is to survey how fairness is formalized in the machine learning literature for the task of prediction and present these formalizations with their corresponding notions of distributive justice from the social sciences literature.
We provide theoretical as well as empirical critiques of these notions from the social sciences literature and explain how these critiques limit the suitability of the corresponding fairness formalizations to certain domains. We also suggest two notions of distributive justice which address some of these critiques and discuss avenues for prospective fairness formalizations.
\end{abstract}

\section{Introduction}
Discrimination refers to unfavourable treatment of people due to the membership to certain demographic groups that are 
distinguished by the attributes protected by law (henceforth, \textit{protected attributes}). Discrimination, based on many attributes and in several domains, is prohibited by international legislation. 
Nowadays, machine learning algorithms are increasingly being used in high-impact domains such as credit, employment, education, and criminal justice which are prone 
to discrimination. 
The goal of fairness in prediction with machine learning is to design algorithms that make \textit{fair} predictions devoid of discrimination. 

\textbf{The aim of this article is to survey how fairness is \textit{formalized} in the machine learning literature and present these formalizations with their corresponding notions from the social sciences literature.} 
The fairness formalizations in the machine learning literature correspond to the notions of \textit{distributive justice} from the social sciences literature, as we discuss in Section \ref{sec:PastNotions}.
Since, some formalizations of fairness can be conflicting with others, the predictions produced by the algorithms using them would vastly differ as well.
Therefore, from the practical point of view, it is important to study how fairness is formalized in the machine learning literature and the implications of various formalizations. To this end, \textbf{we present theoretical as well as empirical critiques of their corresponding notions from the social sciences literature.} 
The co-presentation is with the intention to assist in \textbf{determining the suitability of the existing formalizations of fairness in machine learning literature} and \textbf{building newer formalizations of fairness}. 
In Section \ref{sec:ProspetiveNotions}, we \textbf{nominate two notions from the social sciences literature} which answer some of the critiques of the existing formalizations in the machine learning literature.
Lastly, in Section \ref{sec:Discussion}, we discuss avenues for prospective fairness
formalizations. We begin by formulating the problem of prediction with machine learning.

\textbf{Mathematical formulation of prediction with machine learning:}
Let $\Popu$, $\protAttr$ and $\unprotAttr$ represent a set of individuals i.e. a \textit{population}, protected attributes and remaining attributes respectively.  
Each of the individuals can be assigned an outcome from a finite set $\SetOutcomes$. 
Some of the prediction outcomes are considered to be more beneficial or desirable than others. 
For an individual $\Indv_i \in \Popu$, let $\Outcome_i$ be the true outcome (or label) to be predicted. A (possibly randomized) predictor can be represented by a mapping $\predictor : \Popu \rightarrow \SetOutcomes$ from population $\Popu$ to the set of outcomes $\SetOutcomes$, such that $\predictor(\Indv_i)$ is the predicted outcome for individual $\Indv_i$. A group-conditional predictor consists of a set of mappings, one for each group of the population, $ \predictor =  \{\predictor_S\} \all S \subset \Popu.$ For the sake of simplicity, assume that the groups induce a partition of the population.



\section{What is fair? (Formalizations of fairness in prediction with machine learning)}
\label{sec:PastNotions}
The first step in formalizing fairness in prediction with machine learning is to answer the following two questions:
\begin{minipage}{\columnwidth}\centering
\captionof{table}{The surveyed formalizations of fairness}
\vspace{0.25cm}
\begin{tabular}{c | c | c}
	& Parity & Preference \\
    \hline
    
    Treatment & Unawareness & Preferred treatment \\
    		  & Counterfactual measures & \\
    \hline
    	    & Group fairness &  \\
    Impact	& Individual fairness & Preferred impact \\
            & Equality of opportunity & \\
	
\end{tabular}
\label{tab:AllNotions}
\end{minipage}
\begin{itemize}
\item \textbf{Parity or preference?} : whether fairness means achieving parity or satisfying the preferences.  
\item \textbf{Treatment or impact?} : whether fairness is to be maintained in treatment or impact (results).
\end{itemize}
Next, we will see the existing formalizations of fairness in the machine learning literature. Table \ref{tab:AllNotions} summarizes how they answer the questions presented above.  


\subsection{Fairness through unawareness}
Any predictor which is not group-conditional satisfies this measure. Formally, it is defined as follows:
\vspace{-0.2cm}
\begin{myDefinition}(\textnormal{fairness through unawareness})
A predictor is said to achieve fairness through unawareness if protected attributes are not explicitly used in the prediction process. 
\end{myDefinition}
\vspace{-0.2cm}
 A number of proposed predictors in the machine learning literature satisfy this measure \cite{Dwork:2012:FTA:2090236.2090255, Luong:2011:KIS:2020408.2020488}, while some don't \cite{Calders:2010:TNB:1842547.1842562, NIPS2016_6374, Kamishima:2012:FCP:2405742.2405746}. However, satisfying fairness through unawareness is not a sufficient condition to avoid discrimination when other background knowledge is available \cite{Pedreshi:2008:DDM:1401890.1401959}.
Furthermore, some of the assumptions made during the construction of a predictor might not hold in real-life scenarios \cite{Calders2013} which leads to discrimination even while satisfying this measure.

From the point of view of distributive justice, fairness through unawareness corresponds to the approach of being ``blind'' to counter discrimination.
However, various discriminatory practices have been documented following race-blind approach in education, housing, credit, criminal justice system \cite{BonillaSilva2013, Taslitz2007}.  
It has shown that, in the long run, race-blind approach is less efficient than race-conscious approach \cite{Fryer2008}. Alternatively, some studies show that a blind approach can  work for some specific tasks \cite{Goldin2000}.

The above critiques challenge the suitability of fairness through unawareness to domains in which,
protected attributes can be deduced from easily available non-protected attributes and structural barriers, which hinder the protected groups, are shown to be present by credible surveys. 


\subsection{Counterfactual measures}
These measures model fairness through tools from causal interference. \citet{NIPS2017_6995} recently introduced one such measure which can be defined as follows: 
\vspace{-0.2cm}
\begin{myDefinition}
A predictor $\predictor$ is counterfactually fair, given  $\unprotAttr = z$ and $\protAttr = a$, for all $\Outcome$ and $a \neq a'$, iff \vspace{0.15cm}\\
$
\bP \{ \predictor_{\protAttr = a} = \Outcome \given \unprotAttr = z, \protAttr =a\} =\bP \{ \predictor_{\protAttr = a'} = \Outcome \given \unprotAttr = z, \protAttr =a\}
$ 

\end{myDefinition}
\vspace{-0.2cm}
In the above definition, $\predictor_{\protAttr = a}$ is to be interpreted as the outcome of the predictor $\predictor$ if $\protAttr$ had taken value $a$. For the mathematical details of how such a statement is realized, refer to \citet{NIPS2017_6995}. This measure deems a predictor to be fair if its output remains the same when the protected attribute is flipped to its counterfactual value.      
This measure  compares every individual with a different version of themselves. A similar measure was introduced independently by \citet{NIPS2017_6668}. 

In the literature of social sciences, the closest correspondent to these measures is the theory for counterfactual reasoning given by \citet{Lewis1973-LEWC}. There has been research to indicate that counterfactual reasoning is susceptible to hindsight bias \cite{PETROCELLI201061, 80e08f5f44b24ace9987ca4bdb798187} and outcome bias (i.e. evaluating the quality of a decision when its outcome is already known) \cite{Baron1988}. Moreover, it has been argued that counterfactual reasoning may negatively influence the process of identifying causality \cite{ROESE19971, doi:10.1080/01621459.2000.10474210}.    

These critiques bring into question the suitability of counterfactual measures for potential domains for prediction using machine learning like health-care system or judicial system where the above-mentioned biases are frequently observed. 

\subsection{Group fairness (Statistical/demographic parity)}
Group fairness imposes the condition that the predictor should predict a particular outcome for individuals across groups with \textit{almost} equal probability. 
\vspace{-0.2cm}
\begin{myDefinition}(\textnormal{Group fairness})
A predictor $\predictor : \Popu \rightarrow \SetOutcomes$ achieves group fairness with bias $\epsilon$ with respect to groups $S, T \subseteq \Popu$ and $O \subseteq A$ being any subset of outcomes iff \vspace{0.15cm}\\
$
| \bP \{ \predictor(\Indv_i) \in O \given \Indv_i \in S \} - \bP \{\predictor(\Indv_j) \in O \given \Indv_j \in T \} | \leq \epsilon
$
\end{myDefinition}
\vspace{-0.2cm}
From the above definition it is clear that, group fairness imposes the condition of statistical and demographic parity on the predictor.
Unlike some of the other formalizations of fairness, group fairness is independent of the ``ground truth'' i.e. the label information.  
This is useful when reliable ground truth information is not available e.g. 
in domains like employment, housing, credit and criminal justice, discrimination against protected groups has been well-documented \cite{Pager2008, Waddell2016}. 
Alternatively, in the cases where disproportionality in the respective outcomes can be justified by using non-protected attributes (which don't merely serve as a proxy for protected attributes), imposing statistical parity leads to incorrect outcomes 
and may amount to discrimination against qualified candidates \cite{Luong:2011:KIS:2020408.2020488}. 
Another deficiency of group fairness is that the predictor is not stipulated to select the most ``qualified'' individuals within the groups as long as it maintains 
statistical parity \cite{Dwork:2012:FTA:2090236.2090255}.



The formalization of group fairness follows from the notion of \textit{collectivist egalitarianism} for distributive justice. 
In practice, the biggest (in terms of the number of people affected) implementation of group fairness is the application of affirmative action \cite{DeshpandeBook2013} in India and USA to address discrimination on the basis of caste \cite{Dumont1980}, race and gender. 
See \citet{Weisskopf2004} for arguments made for and against affirmative action polices in both India and the USA. 
Two of the standard objections to group fairness are: it is not meritocratic and it reduces efficiency. 

The underlying assumption behind the first claim is that the allocation of social benefits without affirmative action is meritocratic. However, several studies \cite{DeshpandeA2007, 10.1257/0002828042002561, 147186} have confirmed discrimination on the basis of protected attributes. 
For the second claim, \citet{Holzer2000} conclude on the basis of several studies that ``the empirical case against Affirmative Action on the grounds of efficiency is weak at best''.  
In India, a study by \citet{Deshpande2016} found no evidence of loss in efficiency because of affirmative action policies. Nonetheless, deficiencies mentioned earlier limit the applicability of group fairness. 

\subsection{Individual fairness}
Individual fairness ascertains that a predictor is fair if it produces similar outputs for similar individuals. 
\vspace{-0.2cm}
\begin{myDefinition}(\textnormal{Individual fairness})
\label{def:IndFairness}
A predictor achieves individual fairness iff 
$ \predictor(\Indv_i) \approx \predictor(\Indv_j) \given  \IndDistanceMetric(\Indv_i , \Indv_i) \approx 0 $
where $\IndDistanceMetric : \Popu \times \Popu \rightarrow \mathbb{R}$ is a distance metric for individuals. 
\end{myDefinition}
\vspace{-0.2cm}
Several works including \citet{Dwork:2012:FTA:2090236.2090255} and \citet{Luong:2011:KIS:2020408.2020488} use this notion of fairness. The notion of individual fairness can be then captured by $(\DistDistanceMetric, \IndDistanceMetric)$-Lipschitz property which states that 
$ \DistDistanceMetric(\DistOutcomes(\Indv_i)_\SetOutcomes, \DistOutcomes(\Indv_j)_\SetOutcomes) \leq \IndDistanceMetric(\Indv_i, \Indv_j) $ where $\DistDistanceMetric$ is a distance measure for distributions. Furthermore, \citet{Dwork:2012:FTA:2090236.2090255} prove that if a predictor satisfies $(\DistDistanceMetric, \IndDistanceMetric)$-Lipschitz property, then it also achieves statistical parity with certain bias. 

In the social sciences literature, this formalization is equivalent to \textit{individualism egalitarianism}. 
According to \citet{doi:10.1086/291523}, this is the formal principle of justice. 
This notion delegates the responsibility of ensuring fairness from the predictor to the distance metric. If the distance metric uses the protected attributes directly or indirectly to compute the distance between two individuals, a predictor satisfying Definition \ref{def:IndFairness} could still be discriminatory. Therefore, the potency of this notion of fairness to prevent discrimination depends largely upon the distance metric used. Hence, individual fairness as stated above, can not be considered suitable for domains where reliable and non-discriminating distance metric is not available \footnote{ \citet{Dwork:2012:FTA:2090236.2090255} have provided some approaches to build distance metrics.}.

\subsection{Equality of opportunity}
In the literature of machine learning, the formalization of equality of opportunity was introduced by \citet{NIPS2016_6374}. 
An equivalent formalization was also proposed concurrently and independently by \citet{Zafar:2017:FBD:3038912.3052660}. To formalize it, let us consider the case of binary outcomes with a single beneficial outcome $y=1$.
\vspace{-0.25cm}
\begin{myDefinition} (\textnormal{Equal opportunity}) A predictor is said to satisfy equal opportunity with respect to group $S$ iff
$
\bP \{ \predictor(\Indv_i) = 1 \given \Outcome_i = 1, \Indv_i \in S \} = \bP \{ \predictor(\Indv_j) = 1 \given \Outcome_j = 1, \Indv_j \in \Popu \setminus S \} 
$.
\end{myDefinition}
\vspace{-0.25cm}
It can be considered as a stipulation which states that the true positive rate should be the same for all the groups. An equivalent notion proposed by \citet{Zafar:2017:FBD:3038912.3052660}, called \textit{disparate mistreatment}, asks for the equivalence of misclassification rates across the groups.    

In the social sciences literature, the corresponding notion was presented by \citet{Rawls1971-RAWATO-4}. 
An essay by \citet{10.2307/4320904} states that equality of opportunity would not be able to cope with the problems of \textit{stunted ambition} and 
\textit{selection by bigotry}.
The notion of equality of opportunity has also been criticized for not considering the effect of discrimination due to protected attributes like gender \cite{Okin1991-OKIJGA} and race \cite{Shiffrin2004}. 
It has been shown that the protected attributes like race and gender affect one's access to opportunities in domains such as education, business, politics in many parts of the world \cite{Iqbal2015}. 
The exclusion of attributes like race and gender from the list of attributes deemed to be affecting an individual's life prospects in the notion of equality of opportunity thus calls into question 
its suitability to the domains in which there exists vast evidence that such attributes do indeed affect one's prospects.

\subsection{Preference-based fairness}
\citet{2017arXiv170700010B} introduce two preference-based formalizations of fairness. In order to provide the definitions for the same, the authors first introduce the notion of \textit{group benefit} which is defined as the expected proportion of individuals in the group for whom the predictor predicts the beneficial outcome. 
Group benefit can also be defined as the expected proportion of individuals from the group who receive the beneficial output for whom the true label is the same.
Based on the above notion of group benefit, \citet{2017arXiv170700010B} provide following two fairness formalizations.
\vspace{-0.25cm}
\begin{myDefinition}(\textnormal{Preferred treatment})
A group-conditional predictor is said to satisfy preferred treatment if each group of the population receives more benefit from their respective predictor then they would have received from any other predictor i.e. \vspace{0.15cm}\\
$\Benefit_S(\predictor_S) \geq \Benefit_S(\predictor_T) \smallskip \qquad \all S, T \subset \Popu $

\end{myDefinition}
\vspace{-0.35cm}
\begin{myDefinition}(\textnormal{Preferred impact})
A predictor $\predictor$ is said to have preferred impact as compared to another predictor $\predictor'$ if $\predictor$ offers at-least as much benefit as $\predictor'$ for all the groups. \vspace{0.15cm}\\
$ \Benefit_S(\predictor) \geq  \Benefit_S(\predictor') \qquad  \all S \subset \Popu  $
\end{myDefinition}
\vspace{-0.25cm}
If a classifier is not group-conditional then, it by default satisfies preferred treatment.
In certain applications, there might not be a single universally accepted beneficial outcome. It is possible that a few individuals from a group may prefer another outcome than the one preferred by the majority of the group. In order to alleviate their concerns, the collectivist definition of group benefit needs to be extended to account for individual preferences. 

In the social sciences literature, the above notion corresponds to \textit{envy-freeness} \cite{RePEc:bla:jecsur:v:8:y:1994:i:2:p:155-86}. 
This notion of fairness is attractive because it can be defined in terms of ordinal preference relations of the utility values of the predictors. On the other hand, 
\citet{10.2307/1061111} show that freedom from envy is neither necessary nor sufficient for fairness. 
For many real-world problems, one needs to find fair and efficient solutions amongst the groups. An efficient solution ensures the greatest possible benefit to the groups. In decision making problems, like the domain applications of prediction with machine learning, it can be formally expressed by the notion of \textit{Pareto-efficiency}. 
However, deciding whether there is a Pareto-efficient envy-free allocation is computationally very hard even with simple additive preferences \cite{deKeijzer2009}. 

These critiques indicate that the suitability of such envy-free formalizations is limited only to the domains where such an effective and envy-free allocation can be computed easily.

\section{Prospective notions of fairness}
\label{sec:ProspetiveNotions}
In this section, we describe two prospective notions of fairness which have not been considered in the literature of machine learning so far. Our intent is to address the critique that many of the past formalizations, as seen in Section \ref{sec:PastNotions}, do not offset for the fact that social benefits are being allocated unequally by the algorithms among the people owing to the attributes they had no say in.
\begin{itemize}
\item \textbf{Equality of resources:}
\citet{Dworkin1981-DWOWIE-2} propose the notion of \textit{equality of resources} in which unequal distribution of social benefits is only considered fair when it results from the intentional decisions and actions of the concerned individuals. Equality of resources is \textit{ambition-sensitive} i.e. each individual's ambitions and choices that follow them ascertains the benefits they receive and \textit{endowment-insensitive} i.e. each individual's unchosen circumstances including the natural endowments should be offset. In the second property, equality of resources differs from equality of opportunity as the latter considers differences in natural endowments (including the protected attributes such as sex) as facts of nature which need not be adjusted to achieve fairness. 
\vspace{2mm}
\item \textbf{Equality of capability of functioning:}
\citet{18084} extends the insight that people should not be held responsible for attributes they had no say in to include personal attributes which cause difficulty in developing \textit{functionings}. Functionings are states of ``being and doing'', that is, various states of existence and activities that an individual can undertake. 
\citet{18603, 18084} argue that variations related to the protected attributes like age, sex, gender, race, caste give individuals unequal powers to achieve goals even when they have the same opportunities. In order to equalize capabilities, people should be compensated for their unequal powers to convert opportunities into functionings. To this point, it sounds similar to quality of resources described above. Crucially however, the notion of equality of capability calls for addressing inequalities due to social endowments (e.g. gender) as well as natural endowments (e.g. sex)
, in contrast to the equality of resources \cite{POST:POST646}.   
\end{itemize}
One of the main strengths of this notion of fairness that it is flexible which allows it to be developed and applied in many different ways \cite{Alkire2002-ALKVFS}. Indeed, this notion has been used in the foundations of human development paradigm by the United Nations \cite{doi:10.1080/1354570022000077980, FukudaKumar2003}. One of the major criticism of Equality of capability theory concerns the failure to identify of valuable capabilities \cite{doi:10.1080/00220389608422460}. Another criticism is that the informational requirement of this approach can be very high  \cite{Alkire2002-ALKVFS}. The second criticism applies to equality of resources as well and it makes exact mathematical formalizations of these notions a potentially difficult problem. However, the suitability of these prospective formalizations (unlike the current formalizations) to domains in which natural endowments or social endowments or both impede an individual's prospect to receive social benefits makes the open problem of formalizing them worthwhile. We intend this article to serve as a call for machine learning experts to work on formalizing them.   

\section{Discussion and further directions}
\label{sec:Discussion}
As the field of fairness in machine learning prediction algorithms is evolving rapidly, it is important for us to analyze the fairness formalizations considered so far. To this end, we juxtaposed the fairness notions previously considered in the machine learning literature with their corresponding theories of distributive justice in the social sciences literature. We saw the theoretical critique and analysis of these fairness notions from the social sciences literature. 
Such critiques of the formalizations and experimental studies of their use in large-scale practice serve as guiding principles while choosing the fairness formalizations to use in particular domains.  

We also proposed two prospective notions of fairness, which have been studied extensively in the social sciences literature. Of course, we do not claim that these notions will serve as panacea for all the critiques of the current notions. Our intention is to initiate a discussion about fairness formalizations in prediction with machine learning which recognize that - \textbf{the problem of fair prediction cannot be addressed without considering social issues such as unequal access to resources and social conditioning. While these factors are difficult to quantify and formalize mathematically, it is important to acknowledge their impact and attempt to incorporate them in fairness formalizations.}
{\tiny
\bibliographystyle{ACM-Reference-Format}
\bibliography{references} 
}
\end{document}